\documentclass[runningheads]{llncs}
\usepackage{graphicx}
\usepackage{amsmath,amssymb} 
\usepackage{color}

\newcommand{\Rmnum}[1]{\expandafter\@slowromancap\romannumeral #1@}
\newcommand{\ie}{{\it i.e. }}
\newcommand{\eg}{{\it e.g. }}

\newcommand{\Rid}{{\it Re-id}}
\newcommand{\reid}{{\it re-id}}

\def \z{\mathbf{z}}
\def \h{\mathbf{h}}

\usepackage[width=122mm,left=12mm,paperwidth=146mm,height=193mm,top=12mm,paperheight=217mm]{geometry}

\begin{document}
\pagestyle{headings}
\mainmatter
\def\ECCV14SubNumber{***}  

\title{A Novel Visual Word Co-occurrence Model for Person Re-identification} 

%

\author{Ziming Zhang, Yuting Chen, Venkatesh Saligrama}
\institute{Boston University, Boston, MA 02215\\ \{zzhang14, yutingch, srv\}@bu.edu}

\maketitle

\begin{abstract}
Person re-identification aims to maintain the identity of an individual in diverse locations through different non-overlapping camera views. The problem is fundamentally challenging due to appearance variations resulting from differing poses, illumination and configurations of camera views. To deal with these difficulties, we propose a novel visual word co-occurrence model. We first map each pixel of an image to a visual word using a codebook, which is learned in an unsupervised manner. The appearance transformation between camera views is encoded by a co-occurrence matrix of visual word joint distributions in probe and gallery images. Our appearance model naturally accounts for spatial similarities and variations caused by pose, illumination \& configuration change across camera views. Linear SVMs are then trained as classifiers using these co-occurrence descriptors. On the VIPeR \cite{gray2007evaluating} and CUHK Campus \cite{Zhao_ICCV13_salience} benchmark datasets, our method achieves 83.86\% and 85.49\% at rank-15 on the Cumulative Match Characteristic (CMC) curves, and beats the state-of-the-art results by 10.44\% and 22.27\%.

\end{abstract}

\section{Introduction}\label{sec:intr}
In intelligent surveillance systems, {\it person re-identification} ({\it re-id}) is emerging as a key problem. {\Rid} deals with maintaining identities of individuals traversing different cameras. As in the literature we consider {\reid} for two cameras and focus on the problem of matching probe images of individuals in Camera 1 with gallery images from Camera 2. 
The problem is challenging for several reasons. Cameras views are non-overlapping so conventional tracking methods may fail. Illumination, view angles and configurations for different cameras are generally non-consistent, leading to significant appearance variations to the point that features seen in one camera are often distorted or missing in the other. Finer bio-metrics like face and gait thus often become unreliable~\cite{Vezzani:2013:PRS:2543581.2543596}. 

\begin{figure}
\begin{center}
\centerline{\includegraphics[width=0.9\columnwidth]{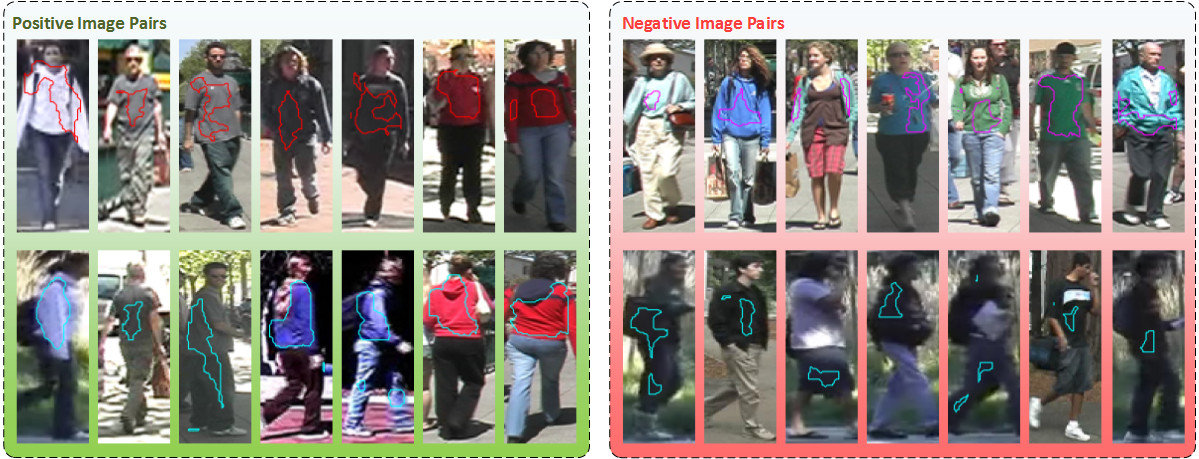}}
\caption{\footnotesize{Illustration of codeword co-occurrence in positive image pairs (\ie two images from different camera views per column belong to a {\em same} person) and negative image pairs (\ie two images from different camera views per column belong to {\em different} persons). For positive (or negative) pairs, in each row the enclosed regions are assigned the same codeword.}}\label{fig:intuition}
\end{center}\vspace{-10mm}
\end{figure}

The existing papers mainly focus on designing distinctive signature to represent a person under different cameras, or learning an effective matching methodology to predict if two images describe the same person. Our proposed method diverts from the literature by aiming to learn an appearance model that is based on {\em co-occurrence statistics} of visual patterns in different camera views. Namely, our appearance model captures the appearance ``transformation'' across cameras instead of some unknown invariant property among different views. Particularly, our method does not assume any smooth appearance transformation across different cameras. Instead, our method learns the visual word co-occurrence pattens statistically in different camera views to predict the identities of persons.

While co-occurrence based statistics has been used in some other works~\cite{DBLP:conf/avss/BanerjeeN11} \cite{Galleguillos2008} \cite{Ladicky:2010:GCB:1888150.1888170}, ours has a different purpose. We are largely motivated by the observation that the co-occurrence patterns of visual codewords behave similar for images from different views. In other words, the transformation of target appearances can be statistically inferred through these co-occurrence patterns. As seen in Fig. \ref{fig:intuition}, we observe that some regions are distributed similarly in images from different views and robustly in the presence of large cross-view variations. These regions provide important discriminant co-occurrence patterns for matching image pairs. For instance, statistically speaking, the first column of positive image pairs shows that ``white'' color in Camera 1 can change to ``light blue'' in Camera 2. However, ``light blue'' in Camera 1 can hardly change to ``black'' in Camera 2, as shown in the first column of negative image pairs.

Thus we propose a novel visual word co-occurrence model to capture such important patterns between images. We first encode images with a sufficiently large codebook to account for different visual patterns. Pixels are then matched into codewords or visual words, and the resulting spatial distribution for each codeword is embedded to a kernel space through {\it kernel mean embedding}~\cite{DBLP:conf/alt/SmolaGSS07} with latent-variable conditional densities \cite{Jebara:2004:PPK:1005332.1016786} as kernels. The fact that we incorporate the spatial distribution of codewords into appearance models provides us with locality sensitive co-occurrence measures. Our approach can be also interpreted as a means to {\em transfer} the information (\eg pose, illumination, and appearance) in image pairs to a common latent space for meaningful comparison. 

To conduct re-identification, we employ linear support vector machines (SVMs) as our classifier trained by the appearance descriptors. On the VIPeR \cite{gray2007evaluating} and CUHK Campus \cite{Zhao_ICCV13_salience} benchmark datasets, our method achieves 83.86\% and 85.49\% at rank-15 on the Cumulative Match Characteristic (CMC) curves, and beats the state-of-the-art results by 10.44\% and 22.27\%.

\subsection{Related Work}

The theme of local features for matching is related to our kernel-based similarity measures. To ensure locality, \cite{Bird:2005:DLI:2218577.2218751} models
the appearances of individuals using features from horizontal strips. \cite{Gheissari_CVPR06_spatiotemporal} clusters pixels into similar groups and the scores are matched based on correspondences of the clustered groups. Histogram features that encode both local and global appearance are proposed in \cite{Bazzani:2012:MPR:2161002.2161235}. Saliency matching \cite{Zhao_ICCV13_salience,Zhao_CVPR13_salience}, one of the-state-of-the-art methods for {\reid} uses patch-level matching to serve as masks in images to localize discriminative patches. More generally low-level features such as color, texture, interest points, co-variance matrices and their combinations have also been proposed \cite{Farenzena_CVPR10_SDALF,Gray_ECCV08_ELF,Prosser_BMVC10_SVR,Bauml_AVSS11_evaluation,Gheissari_CVPR06_spatiotemporal,Bak_AVSS11_MRCG,Ma_BMVC12_Bicov,conf/eccv/LiuGLL12}. In addition high-level structured features that utilize concatenation of low-level features \cite{Ma_BMVC12_Bicov} or deformable part models (DPMs) \cite{Nguyen_NIP13_DPM} have been proposed. Metric learning methods have been proposed for {\reid} (\eg \cite{Dikmen:2010:PRL:1966111.1966152,Li:2012:HRT:2481913.2481917,Mignon_CVPR12_PCCA,Wei-ShiZheng:2011:PRP:2191740.2192190}). In \cite{porikli2003inter,Javed:2008:MIS:1330770.1330930} distance metrics are derived through brightness transfer functions that associate color-levels in the two cameras. \cite{Zheng_PAMI13_RDC} proposes distance metrics that lend importance to features in matched images over the wrongly matched pairs without assuming presence of universally distinctive features. Low-dimensional embeddings using PCA and local FDA have also been proposed \cite{Pedagadi_CVPR13_LFDA}. Supervised methods that select relevant features for {\reid} have been proposed by \cite{Gray_ECCV08_ELF} using Boosting and by \cite{Prosser_BMVC10_SVR} using RankSVMs.

\section{Visual Word Co-occurrence Models}\label{sec:method}
We generally face two issues in visual recognition problems:
(1) {\em visual ambiguity} \cite{DBLP:journals/pami/GemertVSG10} (\ie the appearance of instances which belong to the same thing semantically can vary dramatically in different scenarios), (2) {\em spatial displacement} \cite{DBLP:journals/pami/FelzenszwalbGMR10} of visual patterns. 

While visual ambiguity can be somewhat handled through codebook construction and quantization of images into visual words, our goal of matching humans in {\reid} imposes additional challenges. Humans body parts exhibit distinctive local visual patterns and these patterns systematically change appearance locally. Our goal is to account for this inherent variability in appearance models through co-occurrence matrices that quantify spatial and visual changes in appearance.

\subsection{Locally Sensitive Co-occurrence Designs}
We need co-occurrence models that not only account for the locality of appearance changes but also the random spatial \& visual ambiguity inherent in vision problems. Therefore, we first construct a codebook $\mathcal{Z}=\{\mathbf{z}\}\subset\mathbb{R}^D$ with $M$ codewords. Our codebook construction is global and thus only carries information about distinctive visual patterns. Nevertheless, for a sufficiently large codebook distinctive visual patterns are mapped to different elements of the codebook, which has the effect of preserving local visual patterns. Specifically, we map each pixel at 2D location $\boldsymbol{\pi} \in \Pi$ of image ${\cal I}$ into (at least one) codewords to cluster pixels.

To emphasize local appearance changes, we look at the spatial distribution of each codeword. Concretely, we let $C({\cal I},\z)\subseteq\Pi$ denote the set of pixel locations associated with codeword $\z$ in image ${\cal I}$ and associate a spatial probability distribution, $p(\boldsymbol{\pi}|\mathbf{z},\mathcal{I})$, over this observed collection. In this way visual words are embedded into a family of spatial distributions. Intuitively it should now be clear that we can use the similarity (or distance) of two corresponding spatial distributions to quantify the pairwise relationship between two visual words. This makes sense because our visual words are spatially locally distributed and small distance between spatial distributions implies spatial locality. Together this leads to a model that accounts for local appearance changes. 

While we can quantify the similarity between two distributions in a number of ways, the kernel mean embedding method is particularly convenient for our task. The basic idea to map the distribution, $p$, into a reproducing kernel Hilbert space (RKHS), ${\cal H}$, namely, $p \rightarrow  \mu_p(\cdot) = \sum K(\cdot, \boldsymbol{\pi}) p(\boldsymbol{\pi}) \stackrel{\Delta}{=} E_p(K(\cdot, \boldsymbol{\pi}))$. For universal kernels, such as RBF kernels, this mapping is injective, {\it i.e.}, the mapping preserves the information about the distribution \cite{DBLP:conf/alt/SmolaGSS07}. In addition we can exploit the reproducing property to express inner products in terms of expected values, namely, $\langle \mu_p, \Phi \rangle = E_p(\Phi),\,\forall\, \Phi \in {\cal H}$ and obtain simple expressions for similarity between two distributions (and hence two visual words) because $\mu_p(\cdot) \in {\cal H}$.  

To this end, consider the codeword $\mathbf{z}_m$ in image $\mathcal{I}_i^{(1)}$ and codeword $\mathbf{z}_n$ in image $\mathcal{I}_j^{(2)}$. The co-occurrence matrix (and hence the appearance model) is the inner product of visual words in the RKHS space, namely,
\begin{eqnarray}\label{eqn:kme}
\phi(\mathbf{x}_{ij})_{mn}&=&\left\langle \mu_{p(\cdot \mid \mathbf{z}_m,\mathcal{I}_i^{(1)})}, \mu_{p(\cdot \mid \mathbf{z}_n,\mathcal{I}_j^{(2)})}\right\rangle \nonumber\\
&=& \sum_{\boldsymbol{\pi}_u}\sum_{\boldsymbol{\pi}_v}K(\boldsymbol{\pi}_u,\boldsymbol{\pi}_v)p(\boldsymbol{\pi}_u|\mathbf{z}_m,\mathcal{I}_i^{(1)})p(\boldsymbol{\pi}_v|\mathbf{z}_n,\mathcal{I}_j^{(2)}),
\end{eqnarray}
where we have used the reproducing property in the last equality.
We now have several choices for the kernel $K(\boldsymbol{\pi}_u,\boldsymbol{\pi}_v)$ above. We list some of them here:

\subsubsection{Identity:}
$K(\cdot, \boldsymbol{\pi}) = \mathbf{e}_{\boldsymbol{\pi}}$, where $\mathbf{e}_{\boldsymbol{\pi}}$ is the usual unit vector at location $\boldsymbol{\pi}$. We get the following appearance model:
\begin{equation}\label{eqn:identity}
\phi(\mathbf{x}_{ij})_{mn} \propto \left| C({\cal I}_i^{(1)},\z_m) \bigcap C({\cal I}_j^{(2)},\z_n) \right|,
\end{equation}
where $|\cdot|$ denotes set cardinality. This choice often leads to poor performance in {\reid} because it is not robust to spatial displacements of visual words, which we commonly encounter in {\reid}. 

\subsubsection{Radial Appearance Model (RBF):}
This leads to the following appearance model: 
\begin{eqnarray}\label{eqn:rbf}
\phi(\mathbf{x}_{ij})_{mn} &=& \sum_{\boldsymbol{\pi}_u}\sum_{\boldsymbol{\pi}_v}\exp\left(\frac{\|\boldsymbol{\pi}_u-\boldsymbol{\pi}_v\|_2^2}{2 \sigma^2}\right)p(\boldsymbol{\pi}_u|\mathbf{z}_m,\mathcal{I}_i^{(1)})p(\boldsymbol{\pi}_v|\mathbf{z}_n,\mathcal{I}_j^{(2)})\\
&\leq& \sum_{\boldsymbol{\pi}_u}\max_{\boldsymbol{\pi}_{v}}\left\{\exp\left(\frac{\|\boldsymbol{\pi}_u-\boldsymbol{\pi}_v\|_2^2}{2 \sigma^2}\right)p(\boldsymbol{\pi}_v|\mathbf{z}_n,\mathcal{I}_j^{(2)})\right\}p(\boldsymbol{\pi}_u|\mathbf{z}_m,\mathcal{I}_i^{(1)}).\nonumber
\end{eqnarray}
The upper bound above is used for efficiently computing our appearance model by removing the summation over $\boldsymbol{\pi}_v$. This appearance model is often a better choice than the previous one because RBF accounts for some spatial displacements of visual words for appropriate choice of $\sigma$.

\subsubsection{Latent Spatial Kernel:}
This is a type of probability product kernel that has been previously proposed \cite{Jebara:2004:PPK:1005332.1016786} to encode generative structures into discriminative learning methods. In our context we can view the presence of a codeword $\z_m$ at location $\boldsymbol{\pi}_u$ as a noisy displacement of a true latent location $\h \in \Pi$. The key insight here is that the spatial activation of the two codewords $\z_m$ and $\z_n$ in the two image views ${\cal I}_i^{(1)}$ and ${\cal I}_j^{(2)}$ are conditionally independent when conditioned on the true latent location $\h$, namely,  the joint probability factorizes into $Pr \{ \boldsymbol{\pi}_u,\,\boldsymbol{\pi}_v \mid \h, {\cal I}_i^{(1)}, {\cal I}_j^{(2)}\} = Pr \{ \boldsymbol{\pi}_u \mid \h, {\cal I}_i^{(1)} \}Pr \{ \boldsymbol{\pi}_v \mid \h, {\cal I}_j^{(2)} \}$. We denote the noisy displacement likelihoods, $Pr\{ \boldsymbol{\pi}_u \mid \h, {\cal I}_i^{(1)}\} = \kappa_1(\boldsymbol{\pi}_u,\mathbf{h})$ and $Pr \{ \boldsymbol{\pi}_v \mid \h, {\cal I}_j^{(2)}  \} = \kappa_2 (\boldsymbol{\pi}_v,\mathbf{h})$ for simplicity. This leads us to $K(\boldsymbol{\pi}_u,\boldsymbol{\pi}_v)=\sum_{\mathbf{h}}\kappa_1(\boldsymbol{\pi}_u,\mathbf{h})\kappa_2(\boldsymbol{\pi}_v,\mathbf{h})p(\mathbf{h})$, where $p(\mathbf{h})$ denotes the spatial probability at $\mathbf{h}$, which we assume here to be uniform. By plugging this new $K$ into Eq. \ref{eqn:kme}, we have
\begin{multline}\label{eqn:latent}
\phi(\mathbf{x}_{ij})_{mn} = \sum_{\boldsymbol{\pi}_u}\sum_{\boldsymbol{\pi}_v}\sum_{\mathbf{h}}\kappa_1(\boldsymbol{\pi}_u,\mathbf{h})\kappa_2(\boldsymbol{\pi}_v,\mathbf{h})p(\mathbf{h})p(\boldsymbol{\pi}_u|\mathbf{z}_m,\mathcal{I}_i^{(1)})p(\boldsymbol{\pi}_v|\mathbf{z}_n,\mathcal{I}_j^{(2)}) \\
\leq \sum_{\mathbf{h}}\max_{\boldsymbol{\pi}_u}\left\{\kappa_1(\boldsymbol{\pi}_u,\mathbf{h})p(\boldsymbol{\pi}_u|\mathbf{z}_m,\mathcal{I}_i^{(1)})\right\}\max_{\boldsymbol{\pi}_v}\left\{\kappa_2(\boldsymbol{\pi}_v,\mathbf{h})p(\boldsymbol{\pi}_v|\mathbf{z}_n,\mathcal{I}_j^{(2)})\right\}p(\mathbf{h}),
\end{multline}
where the inequality follows by rearranging the summations and standard upper bounding techniques. 
Again we use an upper bound for computational efficiency, and assume that $\mathcal{P}_{\mathcal{H}}$ is a uniform distribution for simplicity without further learning. The main idea here is that by introducing the latent displacement variables, we have a handle on view-specific distortions observed in the two cameras. We only show the performance using the latent kernel in our experimental section, since it produces much better performance than the other two in our preliminary results.  

\subsection{Implementation of Latent Spatial Kernels}

\begin{figure}[t]
\begin{center}
\centerline{\includegraphics[width=\columnwidth]{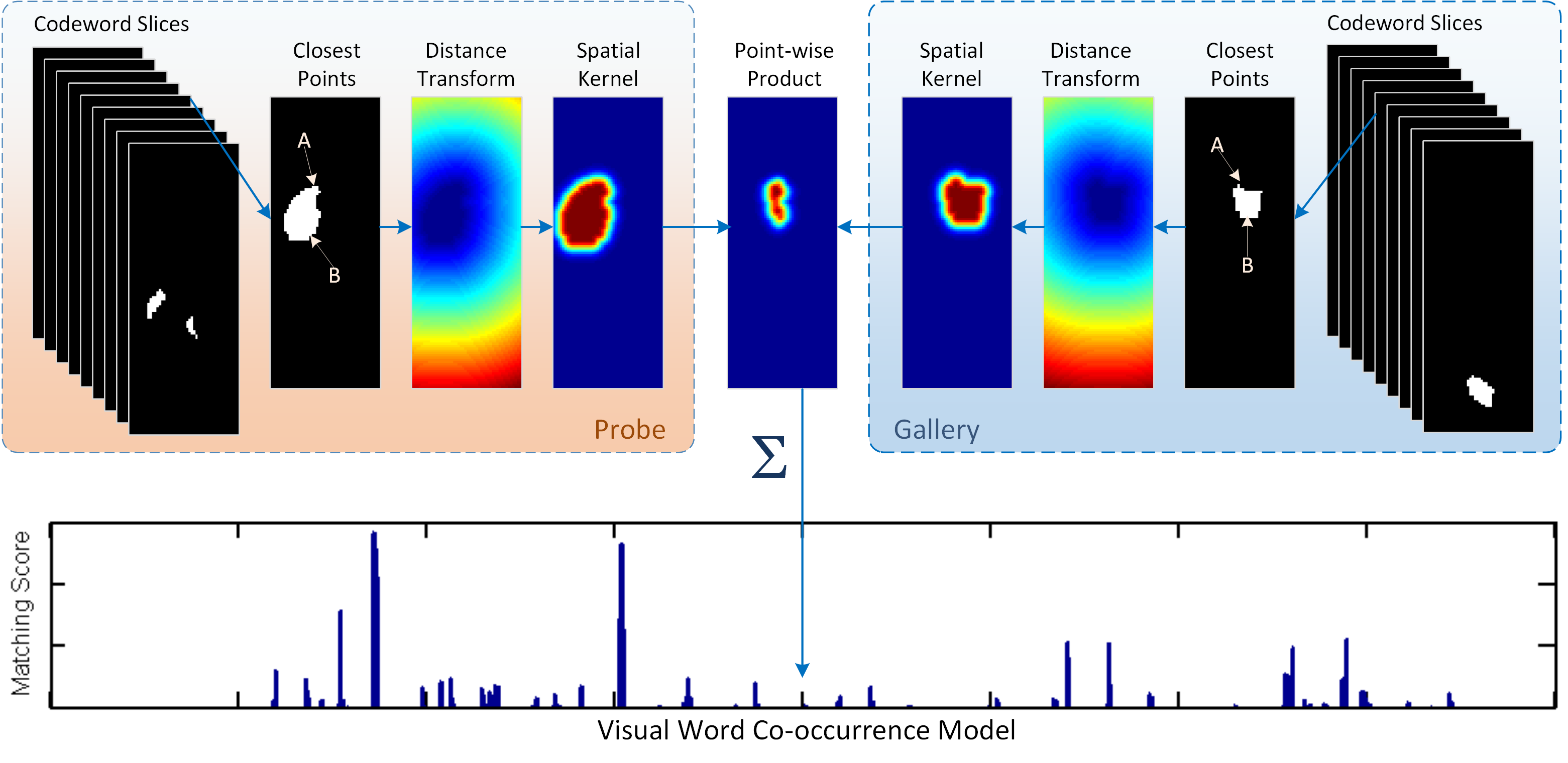}}\vspace{-3mm}
\caption{\footnotesize{Illustration of our visual word co-occurrence model generation process. Here, the white regions in the codeword slices indicate the pixel locations with the same codeword. ``A'' and ``B'' denote two arbitrary pixel locations in the image domain. And ``$\Sigma$'' denotes a sum operation which sums up all the values in the point-wise product matrix into a single value $\phi(\mathbf{x}_{ij})_{mn}$ in our model.}}\label{fig:matching}
\end{center}\vspace{-10mm}
\end{figure}

Fig. \ref{fig:matching} illustrates the whole process of generating the latent spatial kernel based appearance model given the codeword images, each of which is represented as a collection of codeword slices. For each codeword slice, the $\max$ operation is performed at every pixel location to search for the spatially closest codeword in the slice. This procedure forms a distance transform image, which is further mapped to a spatial kernel image. It allows each peak at the presence of a codeword to be propagated smoothly and uniformly. To calculate the matching score for a codeword co-occurrence, the spatial kernel from a probe image and another from a gallery image are multiplied element-wise and then summed over all latent locations. This step guarantees that our descriptor is insensitive to the noise data in the codeword images. This value is a single entry at the bin indexing the codeword co-occurrence in our descriptor for matching the probe and gallery images. As a result, we have generated a high dimensional sparse appearance descriptor.

\section{Experiments}\label{sec:exp}
We test our method on two benchmark datasets, VIPeR \cite{gray2007evaluating} and CUHK Campus \cite{Zhao_ICCV13_salience}. For each dataset, images from separate camera views are split into a gallery set and a probe set. Images from the probe set are treated as queries and compared with every person in the gallery set. For each query, our method produces a ranking of matching individuals in the gallery set. Performance can be evaluated with these resultant rankings, since the identity label of each image is known. The rankings for every possible query is combined into a Cumulative Match Characteristic (CMC) curve, which is a standard metric for re-identification performance. The CMC curve displays an algorithm's recognition rate as a function of rank. For instance, a recognition rate at rank-$r$ on the CMC curve denotes what proportion of queries were correctly matched to a corresponding gallery individual at rank-$r$ or better. Experimental results are reported as the average CMC curve over 3 trials.

\subsection{Implementation}
\begin{figure}[t]
\begin{center}
\centerline{\includegraphics[width=0.9\columnwidth, trim = 0 0 0 25]{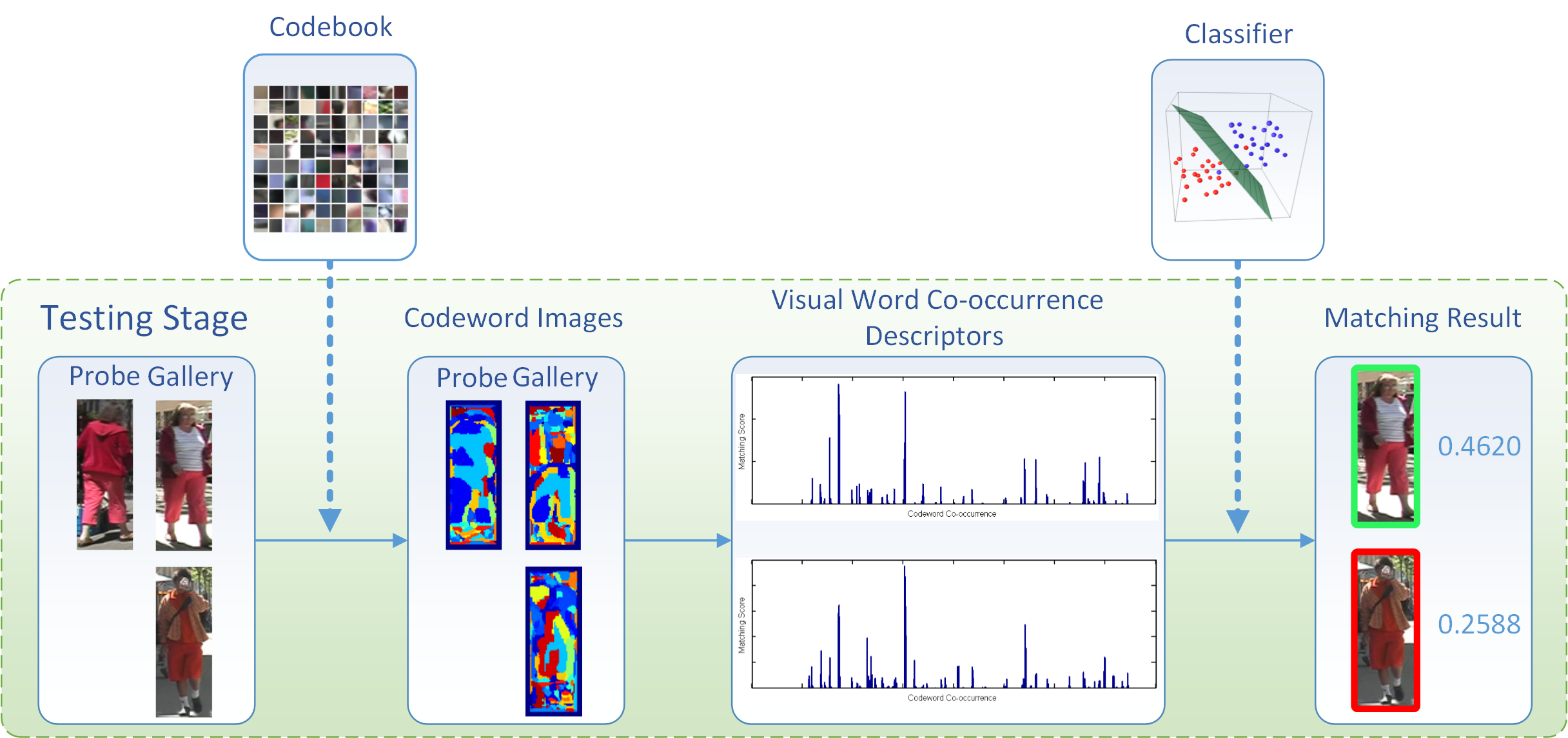}}
\caption{\footnotesize{The pipeline of our method, where ``codebook'' and ``classifier'' are learned using training data, and each color in the codeword images denotes a codeword. This figure is best viewed in color.}}\label{fig:Pipeline}
\end{center}\vspace{-10mm}
\end{figure}

We illustrate the schematics of our method in Fig. \ref{fig:Pipeline}. At training stage, we extract low-level feature vectors from randomly sampled patches in training images, and then cluster them into codewords to form a codebook, which is used to encode every image into a codeword image. Each pixel in a codeword image represents the centroid of a patch that has been mapped to a codeword. Further, a visual word co-occurrence model (descriptor) is calculated for every pair of gallery and probe images, and the descriptors from training data are utilized to train our classifier, performing re-identification on the test data.

Specifically, for each image a 672-dim ColorSIFT \cite{Zhao_ICCV13_salience}\footnote{The authors' code can be downloaded at \url{http://www.ee.cuhk.edu.hk/~rzhao/}.} feature vector is extracted for a 10$\times$10 pixel patch centered at every possible pixel.
Further, we decorrelate each feature using the statistics learned from training data, as suggested in \cite{HariharanMR_ECCV_2012}. 

For codebook construction, we randomly sample 1000 patch features per image in the training set, and cluster these features into a codebook using K-Means. Then we encode each patch feature in images from the probe and gallery sets into a codeword whose Euclidean distance to the patch feature is the minimum among all the codewords. As a result, each image is mapped into a codeword image whose pixels are represented by the indices of the corresponding encoded codewords. We also normalize our appearance descriptors using min-max normalization. The min value is for our descriptors is always 0, and the max value is the maximum among all the codeword co-occurrence bins over every training descriptor. This max value is saved during training and utilized for normalization during testing. 

In the end for classifiers, we employ LIBLINEAR \cite{REF08a}, an efficient linear SVMs solver, with the $\ell_2$ norm regularizer. The trade-off parameter $c$ in LIBLINEAR is set using cross-validation.


\subsection{VIPeR}

Since introduced in \cite{Gray_PETS07_VIPER}, the VIPeR dataset has been utilized by most person re-identification approaches as a benchmark. VIPeR is comprised of 632 different pedestrians captured in two different camera views, denoted by CAM-A and CAM-B, respectively. Many cross-camera image pairs in the dataset have significant variations in illumination, pose, and viewpoint, and each image is normalized to 128$\times$48 pixels.

In order to compare with other person re-identification methods, we followed the experimental set up described in \cite{Zhao_ICCV13_salience}. The dataset is split in half randomly, one partition for training and the other for testing. In addition, samples from CAM-A form the probe set, and samples from CAM-B form the gallery set. The parameter $\sigma$ in the spatial kernel function is set to 3 for this dataset.

\begin{figure}[t]
\begin{minipage}[b]{0.5\linewidth}
 \begin{center}
 \centerline{\includegraphics[width=1.1\columnwidth]{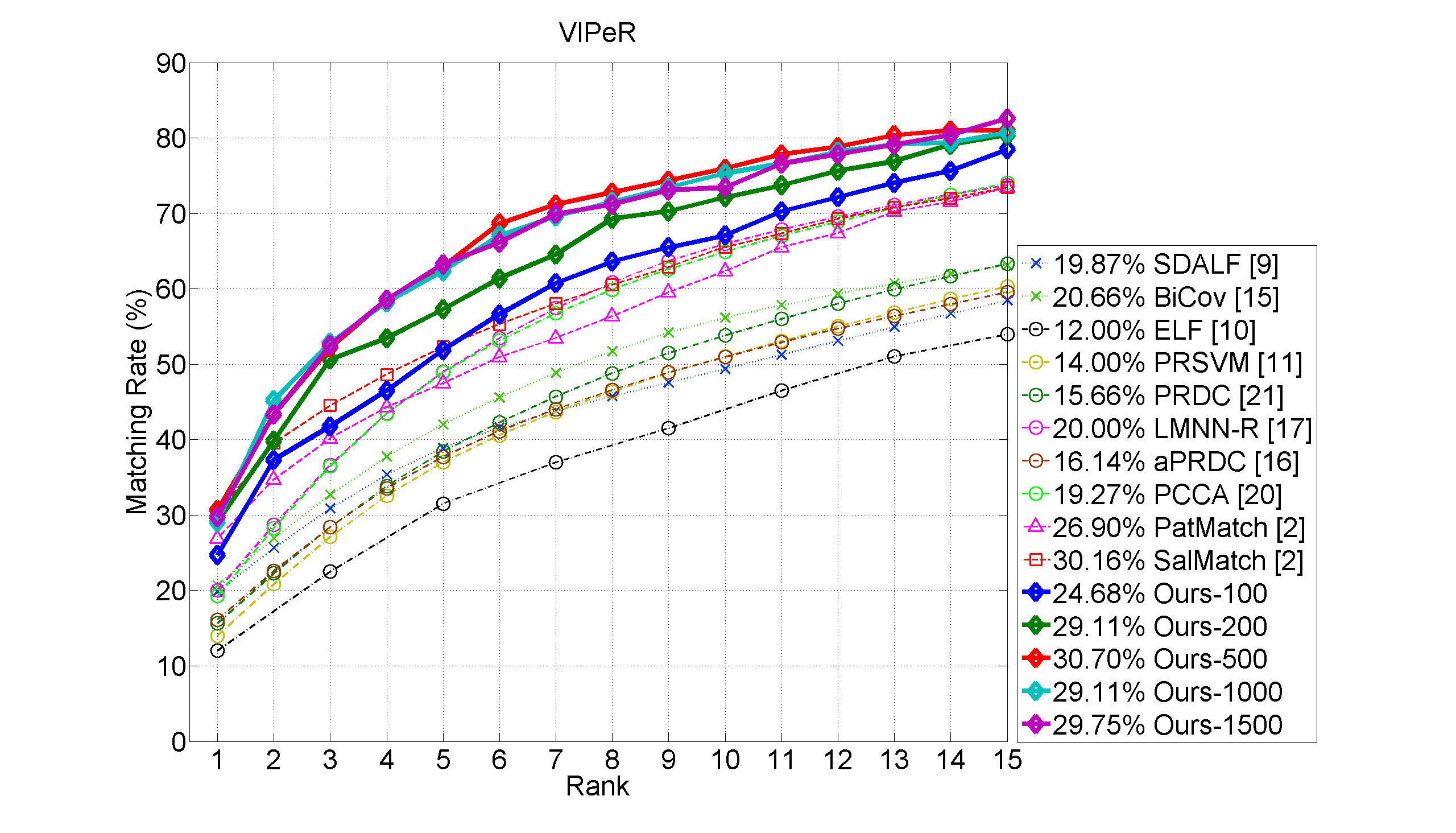}}
 \centerline{\footnotesize{(a) VIPeR}}
 \end{center}
\end{minipage}
\begin{minipage}[b]{0.5\linewidth}
 \begin{center}
 \centerline{\includegraphics[width=1.1\columnwidth]{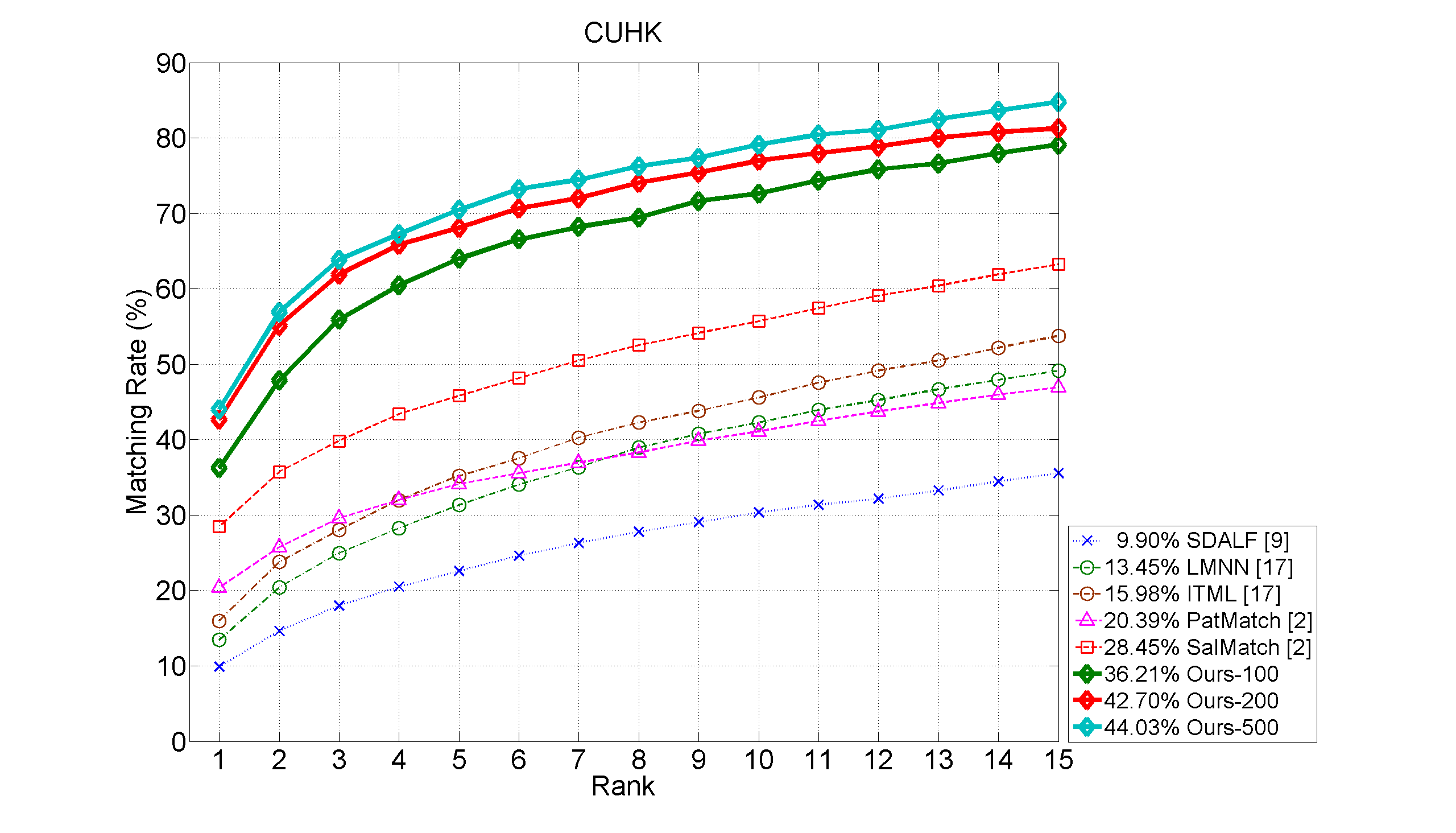}}
 \centerline{\footnotesize{(b) CUHK Campus}}
 \end{center}
\end{minipage}\vspace{-0mm}
\caption{\footnotesize{Matching rate comparison between different methods on (a) VIPeR and (b) CUHK Campus datasets. Numbers following ``Ours-'' in the legends denote the size of the codebook used in each experiment. Expect for our results, the other CMC curves are cited from \cite{Zhao_ICCV13_salience}. This figure is best viewed in color.}}\label{fig:match_rate}
\vspace{-0mm}
\end{figure}

Fig. \ref{fig:match_rate}(a) shows our matching rate comparison with other methods on this dataset. When the codebook size is 100, which is pretty small, our performance is close to that of SalMatch \cite{Zhao_ICCV13_salience}. With increase of the codebook size, our performance is improved significantly, and has outperformed that of SalMatch by large margins. For instance, at rank-15, our best matching rate is 10.44\% higher. Using larger sizes of codebooks, the codeword representation of each image is finer by reducing the quantization error in the feature space. However, it seems that when the codebook size is beyond 500, our performance is saturated. Therefore, in the following experiments, we only test our method using 100/200/500 codewords.

\begin{figure}[t]
\begin{center}
\centerline{\includegraphics[width=\columnwidth]{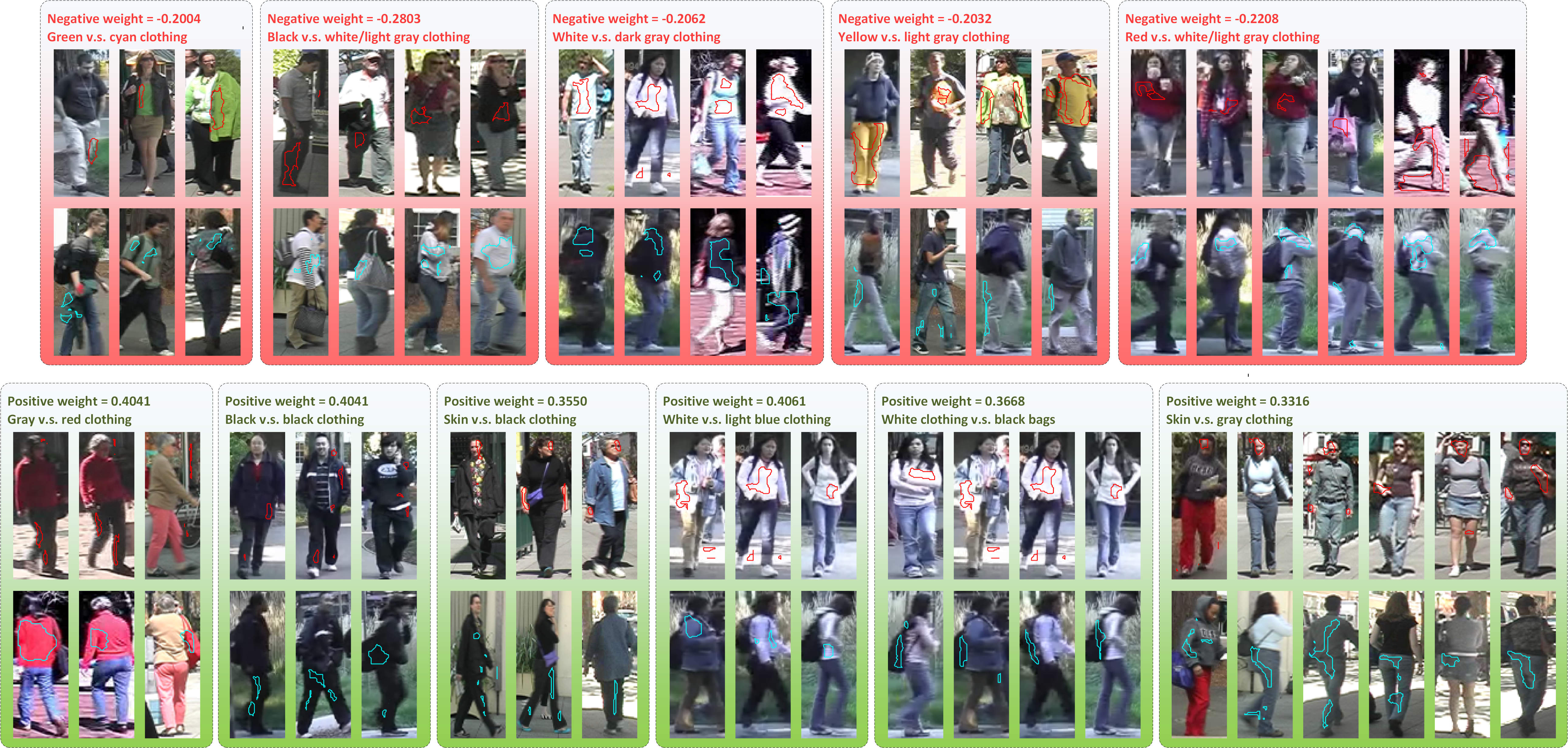}}
\caption{\footnotesize{Examples of codeword co-occurrence with relatively high positive/negative weights in the learned weighting matrix. Same as Fig. \ref{fig:intuition}, in each row the regions enclosed by red (or cyan) color indicate that the codeword per pixel location in these regions is the same. This figure is best viewed in color.}}\label{fig:matching_examples}
\end{center}\vspace{-10mm}
\end{figure}

Fig. \ref{fig:matching_examples} illustrates some codeword co-occurrence examples with relatively high positive/negative weights in the learned weighting matrix. These examples strongly support our intuition of learning codeword co-occurrence based features in Section \ref{sec:intr}.

\subsection{CUHK Campus}

The CUHK Campus dataset is a relatively new person re-identification dataset explored by two state-of-the-art approaches outlined in \cite{Zhao_ICCV13_salience} and \cite{Li_CVPR13_transform}. This dataset consists of 1816 people captured from five different camera pairs, labeled P1 to P5. Each image contains 160$\times$60 pixels. Following the experimental settings from \cite{Zhao_ICCV13_salience} and \cite{Li_CVPR13_transform}, we use only images captured from P1 as our dataset. This subset contains 971 people in two camera views, with two images per view per person. One camera view, which we call CAM-1, captures people either facing towards or away from the camera. The other view, CAM-2, captures the side view of each person.
    
For our experiments, we adopt the settings described in \cite{Zhao_ICCV13_salience} for comparison\footnote{We thank the authors for the response to their experimental settings.}. We randomly select 485 individuals from the dataset and use their 4 images for training, and the rest are used for testing. The gallery and probe sets are formed by CAM-1 and CAM-2, respectively. To re-identify a person, we compare the probe image with every gallery image, leading to 486$\times$2=972 decision scores. Then per person in the gallery set, we average the 2 decision scores belonging to this person as the final score for ranking later. The parameter $\sigma$ in the spatial kernel function is set to 6 for this dataset, since the image size is larger.

Fig. \ref{fig:match_rate}(b) summarizes our matching rate comparison with some other methods. Clearly, using only 100 codewords, our method has already outperformed others dramatically, and it works better when using larger sizes of codebooks, similar to the behavior in Fig. \ref{fig:match_rate}(a). At rank-15, our best performance is 22.27\% better than that of SalMatch.

\section{Conclusion}\label{sec:con}
In this paper, we propose a novel visual word co-occurrence model for person re-identification. The intuition behind our model is that the codeword co-occurrence patterns behave similarly and consistently in pairs of gallery/probe images and robustly to the changes in images. To generate our descriptor, each image is mapped to a codeword image, and the spatial distribution for each codeword is embedded to a kernel space through {\it kernel mean embedding} with latent spatial kernels. To conduct re-identification, we employ linear SVMs as our classifier trained by the descriptors. We test our method on two benchmark datasets, VIPeR and CUHK Campus. On both datasets, our method consistently outperforms other methods. At rank-15, our method achieves matching rates of 83.86\% and 85.49\%, respectively, which are significantly better than the state-of-the-art results by 10.44\% and 22.27\%.

Several questions will be considered as our future work. It would be useful to reduce the computational complexity of calculating our pair-wise latent spatial kernels. One possibility is to modify the learning algorithm by decomposing the weight matrix into two separable parameters, because our appearance model can be decomposed into two parts, one from the probe image and the other from the gallery image. Such decomposition will accelerate the computation. Second, in our preliminary experiments, latent spatial kernel yields significantly better results over the other two choices. It would be interesting to explore other selection of kernels (or even learn the optimal kernels) and how they affect the behavior of our visual word co-occurrence model. Building a {\it re-id} system for natural images using object proposal algorithms (\eg \cite{zhang2011proposal,objectnessBING}) and our model with different classifiers (\eg \cite{zhang2010adamkl,zhang2012efficient,zhang2011learning}) would be interesting as well.

\section*{Acknowledgment}
This material is based upon work supported by the U.S. Department of Homeland Security, Science and Technology Directorate, Office of University Programs, under Grant Award 2013-ST-061-ED0001. The views and conclusions contained in this document are those of the authors and should not be interpreted as necessarily representing the official policies, either expressed or implied, of the U.S. Department of Homeland Security.

\clearpage

\bibliographystyle{splncs}
\bibliography{re-ID}

\begin{thebibliography}{10}

\bibitem{gray2007evaluating}
Gray, D., Brennan, S., Tao, H.:
\newblock Evaluating appearance models for recognition, reacquisition, and
  tracking.
\newblock In: 10th IEEE International Workshop on Performance Evaluation of
  Tracking and Surveillance (PETS). (Sep 2007)

\bibitem{Zhao_ICCV13_salience}
Zhao, R., Ouyang, W., Wang, X.:
\newblock Person re-identification by salience matching.
\newblock In: ICCV. (2013)

\bibitem{Vezzani:2013:PRS:2543581.2543596}
Vezzani, R., Baltieri, D., Cucchiara, R.:
\newblock People reidentification in surveillance and forensics: A survey.
\newblock ACM Comput. Surv. \textbf{46}(2) (December 2013)  29:1--29:37

\bibitem{DBLP:conf/avss/BanerjeeN11}
Banerjee, P., Nevatia, R.:
\newblock Learning neighborhood cooccurrence statistics of sparse features for
  human activity recognition.
\newblock In: AVSS. (2011)  212--217

\bibitem{Galleguillos2008}
Galleguillos, C., Rabinovich, A., Belongie, S.:
\newblock Object categorization using co-occurrence, location and appearance.
\newblock In: CVPR. (June 2008)

\bibitem{Ladicky:2010:GCB:1888150.1888170}
Ladicky, L., Russell, C., Kohli, P., Torr, P.H.S.:
\newblock Graph cut based inference with co-occurrence statistics.
\newblock In: ECCV. (2010)  239--253

\bibitem{DBLP:conf/alt/SmolaGSS07}
Smola, A.J., Gretton, A., Song, L., Sch{\"o}lkopf, B.:
\newblock A hilbert space embedding for distributions.
\newblock In: ALT. (2007)  13--31

\bibitem{Jebara:2004:PPK:1005332.1016786}
Jebara, T., Kondor, R., Howard, A.:
\newblock Probability product kernels.
\newblock JMLR \textbf{5} (December 2004)  819--844

\bibitem{Bird:2005:DLI:2218577.2218751}
Bird, N.D., Masoud, O., Papanikolopoulos, N.P., Isaacs, A.:
\newblock Detection of loitering individuals in public transportation areas.
\newblock Trans. Intell. Transport. Sys. \textbf{6}(2) (June 2005)  167--177

\bibitem{Gheissari_CVPR06_spatiotemporal}
Gheissari, N., Sebastian, T.B., Hartley, R.:
\newblock Person reidentification using spatiotemporal appearance.
\newblock In: CVPR. Volume~2. (2006)  1528--1535

\bibitem{Bazzani:2012:MPR:2161002.2161235}
Bazzani, L., Cristani, M., Perina, A., Murino, V.:
\newblock Multiple-shot person re-identification by chromatic and epitomic
  analyses.
\newblock Pattern Recogn. Lett. \textbf{33}(7) (May 2012)  898--903

\bibitem{Zhao_CVPR13_salience}
Zhao, R., Ouyang, W., Wang, X.:
\newblock Unsupervised salience learning for person re-identification.
\newblock In: CVPR. (2013)  3586--3593

\bibitem{Farenzena_CVPR10_SDALF}
Farenzena, M., Bazzani, L., Perina, A., Murino, V., Cristani, M.:
\newblock Person re-identification by symmetry-driven accumulation of local
  features.
\newblock In: CVPR. (2010)  2360--2367

\bibitem{Gray_ECCV08_ELF}
Gray, D., Tao, H.:
\newblock Viewpoint invariant pedestrian recognition with an ensemble of
  localized features.
\newblock In: ECCV.
\newblock (2008)  262--275

\bibitem{Prosser_BMVC10_SVR}
Prosser, B., Zheng, W.S., Gong, S., Xiang, T., Mary, Q.:
\newblock Person re-identification by support vector ranking.
\newblock In: BMVC. Volume~1. (2010) ~5

\bibitem{Bauml_AVSS11_evaluation}
Bauml, M., Stiefelhagen, R.:
\newblock Evaluation of local features for person re-identification in image
  sequences.
\newblock In: AVSS. (2011)  291--296

\bibitem{Bak_AVSS11_MRCG}
Bak, S., Corvee, E., Bremond, F., Thonnat, M.:
\newblock Multiple-shot human re-identification by mean riemannian covariance
  grid.
\newblock In: AVSS. (2011)  179--184

\bibitem{Ma_BMVC12_Bicov}
Ma, B., Su, Y., Jurie, F.:
\newblock Bicov: a novel image representation for person re-identification and
  face verification.
\newblock In: BMVC. (2012)

\bibitem{conf/eccv/LiuGLL12}
Liu, C., Gong, S., Loy, C.C., Lin, X.:
\newblock Person re-identification: What features are important?
\newblock In: ECCV Workshops (1). Volume 7583. (2012)  391--401

\bibitem{Nguyen_NIP13_DPM}
Nguyen, V.H., Nguyen, K., Le, D.D., Duong, D.A., Satoh, S.:
\newblock Person re-identification using deformable part models.
\newblock In: ICONIP. (2013)  616--623

\bibitem{Dikmen:2010:PRL:1966111.1966152}
Dikmen, M., Akbas, E., Huang, T.S., Ahuja, N.:
\newblock Pedestrian recognition with a learned metric.
\newblock In: ACCV. (2011)  501--512

\bibitem{Li:2012:HRT:2481913.2481917}
Li, W., Zhao, R., Wang, X.:
\newblock Human reidentification with transferred metric learning.
\newblock In: ACCV. (2012)  31--44

\bibitem{Mignon_CVPR12_PCCA}
Mignon, A., Jurie, F.:
\newblock {PCCA:} a new approach for distance learning from sparse pairwise
  constraints.
\newblock In: CVPR. (2012)  2666--2672

\bibitem{Wei-ShiZheng:2011:PRP:2191740.2192190}
Zheng, W.S., Gong, S., Xiang, T.:
\newblock Person re-identification by probabilistic relative distance
  comparison.
\newblock In: CVPR. (2011)  649--656

\bibitem{porikli2003inter}
Porikli, F.:
\newblock Inter-camera color calibration by correlation model function.
\newblock In: ICIP. Volume~2. (2003)  II--133

\bibitem{Javed:2008:MIS:1330770.1330930}
Javed, O., Shafique, K., Rasheed, Z., Shah, M.:
\newblock Modeling inter-camera space-time and appearance relationships for
  tracking across non-overlapping views.
\newblock Comput. Vis. Image Underst. \textbf{109}(2) (February 2008)  146--162

\bibitem{Zheng_PAMI13_RDC}
Zheng, W.S., Gong, S., Xiang, T.:
\newblock Re-identification by relative distance comparison.
\newblock IEEE TPAMI \textbf{35}(3) (2013)  653--668

\bibitem{Pedagadi_CVPR13_LFDA}
Pedagadi, S., Orwell, J., Velastin, S., Boghossian, B.:
\newblock Local fisher discriminant analysis for pedestrian re-identification.
\newblock In: CVPR. (2013)  3318--3325

\bibitem{DBLP:journals/pami/GemertVSG10}
van Gemert, J., Veenman, C.J., Smeulders, A.W.M., Geusebroek, J.M.:
\newblock Visual word ambiguity.
\newblock IEEE Trans. Pattern Anal. Mach. Intell. \textbf{32}(7) (2010)
  1271--1283

\bibitem{DBLP:journals/pami/FelzenszwalbGMR10}
Felzenszwalb, P.F., Girshick, R.B., McAllester, D.A., Ramanan, D.:
\newblock Object detection with discriminatively trained part-based models.
\newblock TPAMI \textbf{32}(9) (2010)  1627--1645

\bibitem{HariharanMR_ECCV_2012}
Hariharan, B., Malik, J., Ramanan, D.:
\newblock Discriminative decorrelation for clustering and classification.
\newblock In: ECCV. (2012)  459--472

\bibitem{REF08a}
Fan, R.E., Chang, K.W., Hsieh, C.J., Wang, X.R., Lin, C.J.:
\newblock {LIBLINEAR}: A library for large linear classification.
\newblock JMLR \textbf{9} (2008)  1871--1874

\bibitem{Gray_PETS07_VIPER}
Gray, D., Brennan, S., Tao, H.:
\newblock Evaluating appearance models for recognition, reacquisition, and
  tracking.
\newblock In: PETS. (2007)  47--47

\bibitem{Li_CVPR13_transform}
Li, W., Wang, X.:
\newblock Locally aligned feature transforms across views.
\newblock In: CVPR. (Jun 2013)  3594--3601

\bibitem{zhang2011proposal}
Zhang, Z., Warrell, J., Torr, P.H.S.:
\newblock Proposal generation for object detection using cascaded ranking svms.
\newblock In: CVPR, IEEE (2011)  1497--1504

\bibitem{objectnessBING}
Cheng, M.M., Zhang, Z., Lin, W.Y., Torr, P.H.S.:
\newblock Bing: Binarized normed gradients for objectness estimation at 300fps.
\newblock In: CVPR, IEEE (2014)

\bibitem{zhang2010adamkl}
Zhang, Z., Li, Z.N., Drew, M.S.:
\newblock Adamkl: A novel biconvex multiple kernel learning approach.
\newblock In: Pattern Recognition (ICPR), 2010 20th International Conference
  on, IEEE (2010)  2126--2129

\bibitem{zhang2012efficient}
Zhang, Z., Sturgess, P., Sengupta, S., Crook, N., Torr, P.H.:
\newblock Efficient discriminative learning of parametric nearest neighbor
  classifiers.
\newblock In: Computer Vision and Pattern Recognition (CVPR), 2012 IEEE
  Conference on, IEEE (2012)  2232--2239

\bibitem{zhang2011learning}
Zhang, Z., Ladicky, L., Torr, P., Saffari, A.:
\newblock Learning anchor planes for classification.
\newblock In: Advances in Neural Information Processing Systems. (2011)
  1611--1619

\end{thebibliography}
\end{document}